\documentclass[twocolumn]{article}

\usepackage[utf8]{inputenc}
\usepackage{titling}
\usepackage{authblk}
\usepackage{graphicx}
\usepackage{microtype}
\usepackage{subcaption}
\usepackage{tikz}
\usepackage{amsmath}
\usepackage{hyperref}

\renewcommand\Affilfont{\small}
\newenvironment{acknowledgements} {\renewcommand\abstractname{Acknowledgements}\begin{abstract}} {\end{abstract}}
\author[1]{Luis Miguel Vieira da Silva}
\author[1]{Aljosha Köcher}
\author[1]{Alexander Fay} 
\affil[1]{Institute of Automation \protect\\ 
Helmut-Schmidt-University Hamburg, Germany \protect\\ 
e-mail: miguel.vieira@hsu-hh.de}
\title{A Capability and Skill Model for Heterogeneous Autonomous Robots}
\date{}
\begin{document}

\begin{titlingpage}
  \maketitle
  \abstract{
    Teams of heterogeneous autonomous robots become increasingly important due to their facilitation of various complex tasks. For such heterogeneous robots, there is currently no consistent way of describing the functions that each robot provides. In the field of manufacturing, capability and skill modeling is considered a promising approach to semantically model functions provided by different machines. This contribution investigates how to apply and extend capability models from manufacturing to the field of autonomous robots and presents an approach for such a capability model.

    \vspace{10 mm}
  }
\end{titlingpage}

\section{Introduction}
The importance of autonomous mobile robots continues to grow due to their wide range of possible applications. Autonomous mobile robots are used, for example, in industry, agriculture or even in private households. An autonomous mobile robot has the ability to move in an unpredictable and partly unknown environment with little or no human intervention. Accordingly, such robots are used to support difficult tasks and services \cite{Alatise.2020}. 
Developing a single robot that can adapt to all circumstances and perform a wide variety of increasingly complex missions is difficult and expensive \cite{Rizk.2020}. 
Furthermore, some tasks are only suitable for certain robots, so there are different types of autonomous mobile robots such as Unmanned Air Vehicles (UAVs), Unmanned Ground Vehicles (UGVs), or Unmanned Surface Vehicles (USVs) \cite{Rizk.2020}. 
These different robots can work together to perform a variety of complex tasks such as transportation, surveillance or search and rescue missions. Such heterogeneous Multi-Robot Systems (MRS) are typically robust to failures because other robots with redundant functions may still accomplish a mission when a single robot fails. Moreover, the strengths of each robot are used and the disadvantages are compensated for \cite{Tuci.2018}. 

Despite growing demand and clear advantages of teams of Heterogeneous Autonomous Robots (HAuRs), there is a problem with the different ways in which their provided functions are described. 
Each robot has functions to perform its tasks. The description of these functions is typically tailored specifically for an individual robot. In teams of HAuRs, robots of different manufacturers may be used. The different and incompatible descriptions of functions already cause problems within one modality (such as air, ground, or water) but aggreviate when robots for different modalities are to be combined in a mission. 
The need for these HAuRs to work together requires a standardized and well-defined way to describe their functions \cite{IEEE1872}. Such an unambiguous function description significantly facilitates communication between HAuRs and thus improves coordination of activities since there is a common understanding. Finally, powerful techniques such as formal reasoning may be employed to find matching functions and thus enable improved autonomy of HAuRs \cite{Bayat.2016}.

In the area of manufacturing, the concept of a ``team'' of systems that try to jointly fulfill tasks and achieve a certain goal is used in the form of Flexible Manufacturing Systems (FMS), which also enable a quick and simple integration of new machines. Such a \emph{Plug \& Produce} concept is achieved by a formal description of machines and their functions in a capability and skill model that separates the description from the implementation of a function. 
A capability is the description of a function, while a skill is the technology-dependent implementation of a function. Semantic technologies are increasingly used here \cite{FKM+_CapabilitiesandSkillsin_26.04.2022}.

Therefore, such an approach towards semantic capability and skill modeling may be promising also for HAuRs and is thus investigated by answering the following question: \emph{To what extent can semantic capability and skill models be applied or extended to teams of heterogeneous autonomous robots?} 
The research goal is to create a formal description for teams of HAuRs.   

In Section~\ref{sec:requirements} requirements for a capability and skill model for HAuRs are determined. In Section~\ref{sec:relatedwork}, an overview of existing approaches to describing functions both in the field of HAuRs in the field of manufacturing is given and evaluated using the criteria from Section~\ref{sec:requirements}. Section~\ref{sec:model} analyzes the most appropriate capability and skill model from manufacturing for applicability to HAuRs, thus establishing a model for HAuRs. Section~\ref{sec:conclusion} gives a summary, shows open issues and future research questions.

\section{Requirements}
\label{sec:requirements}
Modeling functions of HAuRs is promising from many points of view. In the field of HAuRs, mission planning is an important element in which a certain goal, such as the transportation of a workpiece from a warehouse to a factory, is to be accomplished. Therefore descriptions of functions are needed in order to check to what extent this goal can be accomplished with available HAuRs. 
For this purpose, the goal is decomposed into the functions of the HAuRs and results in subtasks for the HAuRs. On the other hand, executions of the functions via a certain technology are necessary for mission execution. The implementation for the execution of functions is not required for planning. Another important issue in the field of HAuRs is their monitoring. Monitoring is performed at both levels considering the current abstract function that the robot is performing but also considering the current state of the specific execution of the function.

After the need for a model to describe functions of HAuRs has been outlined, requirements for such a model are established. Basic requirements for a function model for HAuRs are both \emph{Reusability} and \emph{Extensibility}. The model should not be designed for a specific use case or task, but should be generally applicable to various different use cases in the field of HAuRs \cite{Bayat.2016}. 
In addition, a future use case or the role of a robot within it may be different from the previously defined ones. Furthermore, HAuRs can be modular so that components of a robot can be exchanged for another application. To provide modularity, the model should not be fixed, but easily extendable to add future relevant elements to the model. If specific aspects are missing for a use case, adding these aspects must be supported. 
To fulfill the requirement of \emph{Reusability} and \emph{Extensibility}, a modeling language is necessary which considers a formal and explicit specification of a common conceptualization.

The description of robot functions must be \emph{manufacturer-independent} by using or creating a function model that is accepted by various interest groups \cite{Bayat.2016}. For this purpose, it is advisable to use standards that represent a consensus among experts and increase the reusability at the same time \cite{Hildebrandt.2020}. Teams of HAuRs usually consist of devices from different manufacturers. A manufacturer-independent model is needed to link information from different manufacturers. 

Another requirement is the \emph{Separation between} the abstract description of functions of a autonomous robot, called \emph{Capabilities}, and the technical solution used to execute described capabilities, called \emph{Skills} \cite{Kocher1}. The separation originates from manufacturing and is transferred to HAuRs. 
Accordingly, a capability of an autonomous robot is executed by means of a skill which represents the actual executable robot function.  
The separation between capabilities and skills can be very advantageous when planning missions. Mission planning is done via capabilities and mission execution via the associated skills, so that planning is not done via a multitude of different skills. Multiple skills can exist for the same capability and every skill could in turn be made available via different skill interfaces. For example, a capability \emph{Fly} of a UAV can be executed via a web service but also via MQTT. 
There may be other technologies for execution, but also other identical robots that offer the same capability, but their possible interfaces for execution are different at least in the invocation or even in the technology. A capability can apply to multiple robots, while a skill is a specific implementation of a capability of a robot, which in turn can be invoked via multiple skill interfaces. Accordingly, a capability can be executed via different skills, which in turn can be invoked via even more different skill interfaces. 
Moreover, the capability model can remain the same, while new skills and skill interfaces may be added for upcoming technologies. However, this is only possible if the requirement of extensibility is met.

In addition to these rather basic requirements, more specific requirements are necessary for the field of HAuRs. Autonomous robots are characterized by the fact that they can travel along paths and change their \emph{Position}. Therefore, a description is necessary to specify position and orientation of a robot in three dimensions.

In the context of HAuRs, a robot operates within a specific \emph{Environment}. An environment varies depending on the application. Based on an environment, a robot can adapt its behavior. For example, no-fly zones may be established that must be considered when executing a capability of a robot. On the other hand, environment perception can be used, for example, for collision avoidance as a capability of a robot \cite{OlivaresAlarcos.2019}. Information about the \emph{Environment} must be included in the model.  

Another requirement is the focus on teams of HAuRs and not only the consideration of individual robots. Only by using a team of HAuRs, many complex scenarios become feasible and efficient. Multiple, simple robots are easier and cheaper to develop than a single powerful, complex robot \cite{Bayat.2016}. 
However, for a team of HAuRs results the requirement of \emph{Communication}. By means of communication between robots, information about invocations of the individual skills and their feedback is exchanged so that, e.g., a robot executes its skill after a successful feedback from another robot. No real-time information about, e.g., obstacles is exchanged. So it is essential that an autonomous robot can transmit information to others and also receive and transform information \cite{OlivaresAlarcos.2019}. 

Not only necessary communication emerges from a team of HAuRs, but also physical \emph{Interaction} in the sense of cooperation or collaboration of HAuRs. HAuRs cooperate with each other in the sense that they strive to achieve a common goal. A task may be too challenging for one robot and must therefore be accomplished jointly by decomposing a task into simpler subtasks that single robots can accomplish. Effective cooperation between HAuRs is achieved through communication and coordination \cite{Bayat.2016}.

\section{Related Work}
\label{sec:relatedwork}
In this section, existing contributions to function description of teams of HAuRs are considered and evaluated based on the presented requirements. This is followed by a consideration and evaluation of existing contributions to function modeling in manufacturing. Both categories mainly consist of contributions that focus on formal models such as ontologies. For this reason, a discussion about the suitability of ontologies is given first.  

\subsection{Suitability of Ontologies}
\label{subsec:ontologies}
According to Studer et al., an ontology is defined as a "formal, explicit specification of a shared conceptualization" \cite{Studer.1998}. Ontologies thus specify a \emph{common} understanding about a certain domain. The specification of concepts of a domain is \emph{formal}, thus machine-readable, and \emph{explicit}, i.e., an unambiguous description of the concepts and their relations \cite{Guarino.2009}. 
The explicit specification additionally enables communication between multiple participants of a system \cite{IEEE1872.2}. 
Ontologies can be combined by one ontology importing other ontologies and using new relations to link the individual ontologies. Thereby, the individual ontologies can be managed separately \cite{Hildebrandt.2020}. In Section~\ref{sec:requirements}, such a modeling language was demanded to enable reusability and extensibility.
Ontologies are further reusable because they represent knowledge about a domain independent of a particular task \cite{Bayat.2016}. Collaboration requires mutually agreed and ideally standardized terms, hence this paper uses standards-based ODPs as discussed in \cite{Hildebrandt.2020}. 
Another important reason why ontologies are well suited for modeling functions is \emph{Reasoning}, which automatically infers implicit knowledge from explicitly modeled knowledge, enabling, for example, planning.
Descriptions of functions can be matched in order to automatically compose functions from others and in turn fulfill required functions \cite{Malakuti.2018}. 

\subsection{Function Modeling for Autonomous Robots}
\label{subsec:RW_modelsForRobots}
One approach to modeling robots is KnowRob \footnote{http://knowrob.org}, which is designed for autonomous service robots. The approach has been presented in several publications. 
First introduced in \cite{Tenorth.2009}, this approach introduces a knowledge representation and processing for use in robot control. The approach has been improved and extended subsequently. The approach aims to pass more abstract instructions to robots, which generate on this basis detailed information necessary for execution \cite{Tenorth.2017}. 
KnowRob uses the DOLCE+DnS Ultralite (DUL) upper ontology and augments it with some basic concepts for robotics, such as capabilities and coarse types of robots. The description remains rather abstract. The knowledge representation is not based on any standard. Functions and their executions have been separated, but no description of invocation interface is given. Furthermore, neither communication nor interaction of robots is described semantically. The concept of a semantic map of the environment is addressed to some extent. 

The project ROSETTA aims at "RObot control for Skilled ExecuTion of Tasks in natural interaction with humans; based on autonomy, cumulative knowledge and learning" \cite{Stenmark.2015}. The focus is on industrial robots. In this project, a number of ontologies for robot functions had been developed until the current ontology, ROSETTA, emerged. The ontology uses the core ontology for robotics and automation (CORA) of the IEEE Standard 1872-2015 \cite{IEEE1872}. 
The focus is on the description of devices and abstract functions, but not the description of interfaces to call functions. The description of functions is limited to the description of function types. The use of a basic structure of functions is not included. The environment, interaction or communication is not considered at all. 

\begin{figure}
    \centering
    \includegraphics[width=1.0\linewidth]{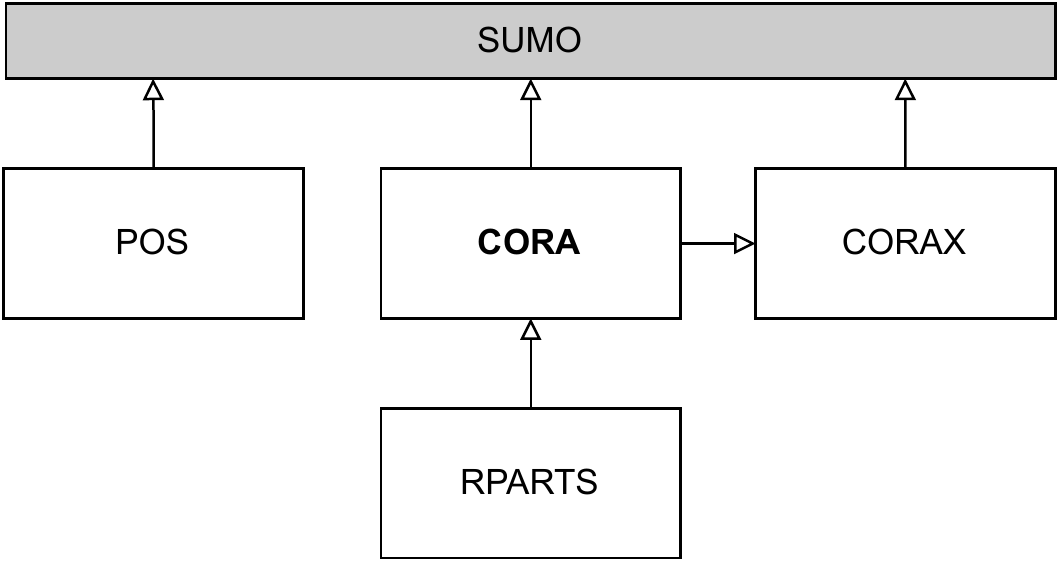}
    \caption{Overview of individual components of the IEEE Standard 1872-2015. CORA is the main component and is linked to the upper ontology SUMO. Likewise, the extensions CORAX, POS and RPARTS are linked to SUMO \cite{Fiorini.2015}.}
    \label{fig:ieee1872}
\end{figure}

In \cite{Stenmark.2015}, a standard for modeling autonomous robots is used, thus \cite{Stenmark.2015} complies with requirement {Reusability} by using standards. The IEEE Standard 1872-2015 \cite{IEEE1872} presents such a standard for ontologies that define key terms in robotics and automation. The ontologies developed there are connected to the Suggested Upper Merged Ontology (SUMO) \cite{Niles.2001}. SUMO defines very general terms like objects or processes. 
CORA is the center of the developed ontologies and defines the three concepts robot, robot group and robot system. In addition, CORAX is used to define concepts such as \emph{Design}, \emph{Physical Environment}, and also POS, which defines concepts related to position and orientation. RPARTS can be used to specify more precisely the types of devices of a robot defined as robot parts in CORA.  Fig.~\ref{fig:ieee1872} provides an overview of the individual components of the standard. 
While \cite{IEEE1872} provides a suitable foundation for developing function modeling, which needs to be extended. 

Therefore, efforts are being made in subgroups such as the Autonomous Robotics (AuR) group to make the ontologies of \cite{IEEE1872} applicable. The AuR group has published the IEEE Standard 1872.2-2021 \cite{IEEE1872.2}, which focuses on HAuRs and is highly relevant to the work of this paper. 
In \cite{IEEE1872.2}, a separation is made between the different modalities of air, water and land. Two upper ontologies, SUMO and DUL, are used here. Environment, interaction, and communication are also considered in \cite{IEEE1872.2}. In particular, in \cite{IEEE1872.2} a clear distinction is made between a behavior of a robot and a function to be performed in order to achieve this behavior. The standard of \cite{IEEE1872.2} is very promising, however, the ontology is not publicly available and only parts are presented in the published standard itself. Furthermore, this approach is still too abstract to be applied in practice.
Neither functions nor their invocation are modeled in detail using existing standards in the field of HAuRs and can be used directly. Due to the lack of availability of the ontology, it is also not entirely clear to what extent the structure of HAuRs in terms of actuators and sensors is considered. Nevertheless, the ontology presented in \cite{IEEE1872.2} serves as a sound foundation for developing applicable function modeling. 

To summarize, in the field of HAuRs, there are only few and insufficient approaches to function modeling. However, the topic of function modeling has been considered for quite some time in the field of manufacturing with robots. Therefore, capability and skill models in manufacturing will be examined in the following sub-section. 

\subsection{Capability and Skill Models in Manufacturing} 
\label{subsec:RW_modelsInManufacturing}
For the domain of manufacturing in the last ten years alone, a wide variety of approaches have been published focusing on a specification and encapsulation of machine functions \cite{FKM+_CapabilitiesandSkillsin_26.04.2022}. 
These approaches mostly aim at making functions available to superordinate systems, e.g., production planning tools, by using a machine-readable description of these functions. 
Goals are to adapt manufacturing systems in an easy manner and to plan and execute manufacturing processes on the basis of these functions.

Over time, various terms have been used for this purpose, e.g., \emph{Service}, \emph{Task}, \emph{Capability} or \emph{Skill}). But in recent years, more and more approaches have been published using the terms \emph{Capabilities} and \emph{Skills}. While these two terms have often been used synonymously in the past, a distinction is increasingly being made. Capabilities are seen as a specification of machine functions, and skills are seen as their executable counterpart, i.e. an implementation of a machine function together with a description of an execution interface \cite{FKM+_CapabilitiesandSkillsin_26.04.2022, KBH+_AReferenceModelfor_15.09.2022b}.

Nevertheless, most research still focuses on either one of these two aspects. Approaches that focus on capabilities typically develop semantic models using ontologies to formally describe machine functions and relevant information such as inputs and outputs. Examples of this group of approaches are \cite{AmDu_AnUpperOntologyfor_2006} and \cite{JSL_FormalResourceandCapability_2016}.
The authors of \cite{AmDu_AnUpperOntologyfor_2006} present Manufacturing Service Description Language, an upper ontology that contains core concepts and relations to describe manufacturing capabilities as manufacturing functions provided by a manufacturer.

Ref. \cite{JSL_FormalResourceandCapability_2016} presents a formal capability model in the form of an ontology that is suited to be used in adaptation of plants. Capabilities are described by their name and parameters. Combined capabilities can be composed of multiple simple ones via so-called capability associations. The presented ontology forms the basis for a matching of provided capabilities with required capabilities. However, it contains no information needed for the execution of skills.

In \cite{WBS+_AnOntologybasedMetamodelfor_9820209112020}, a rather abstract capability meta model is presented which defines generic terms and may thus be used as an upper / domain ontology. The meta model is based on existing ones such as SSN\footnote{https://www.w3.org/TR/vocab-ssn/} and ontologies built on standards such as VDI 2860. Furthermore, patterns for reusing and extending this ontology are shown.
In addition to the meta model terms, \cite{WBS+_AnOntologybasedMetamodelfor_9820209112020} also present a separate robot ontology. 
Elements of the meta model (e.g. the class \emph{Capability}) are referenced in a revised version of the ontology introduced in \cite{Kocher1} --- and are thus used in the ontology presented in this contribution. 
However, the robot ontology of \cite{WBS+_AnOntologybasedMetamodelfor_9820209112020} only contains a rather small taxonomy of robot movement types without ways to specify these movements in detail. Furthermore, it is not properly published online and can thus not be referenced. 

\begin{figure*}[t]
    \centering
    \begin{subfigure}{0.45\textwidth}
    \includegraphics[width=\hsize]{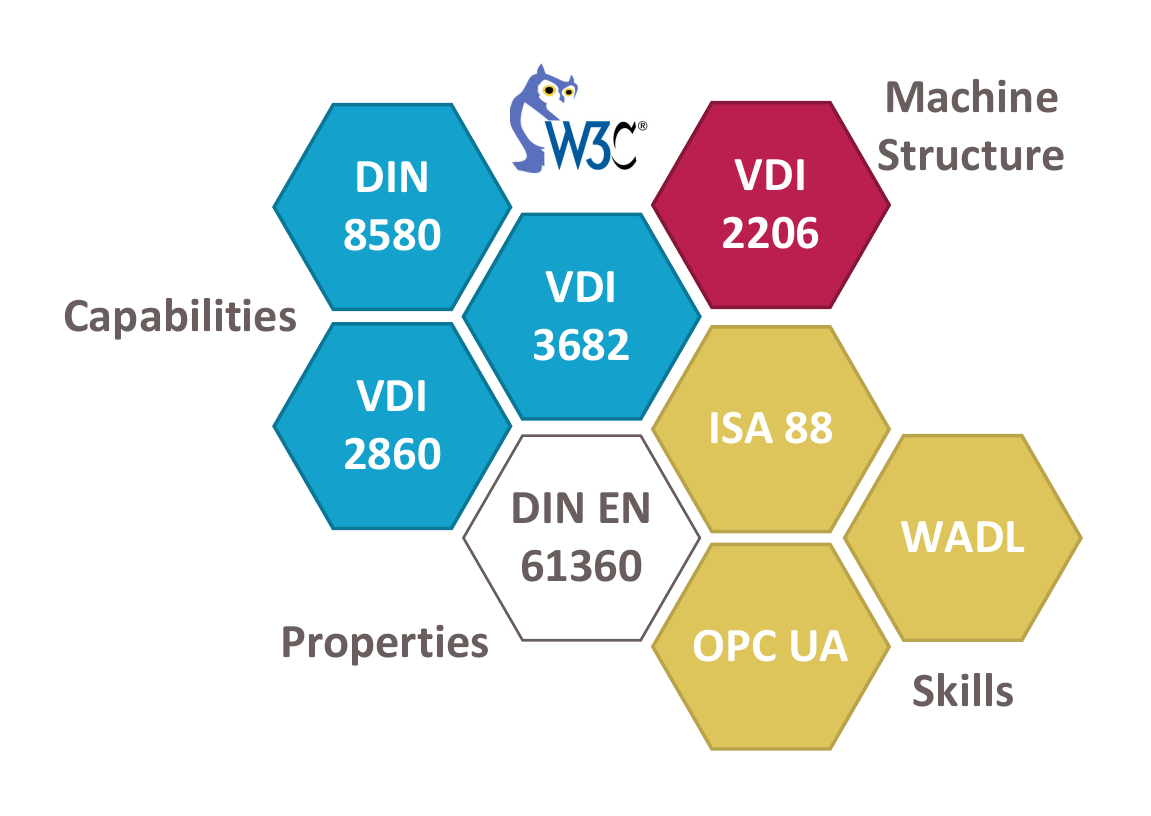}
    \caption{Formal capability and skill model for manufacturing applications \cite{Kocher1}.}
    \label{fig:capability-model-ak}
    \end{subfigure}
    \begin{subfigure}{0.45\textwidth}
    \includegraphics[width=\hsize]{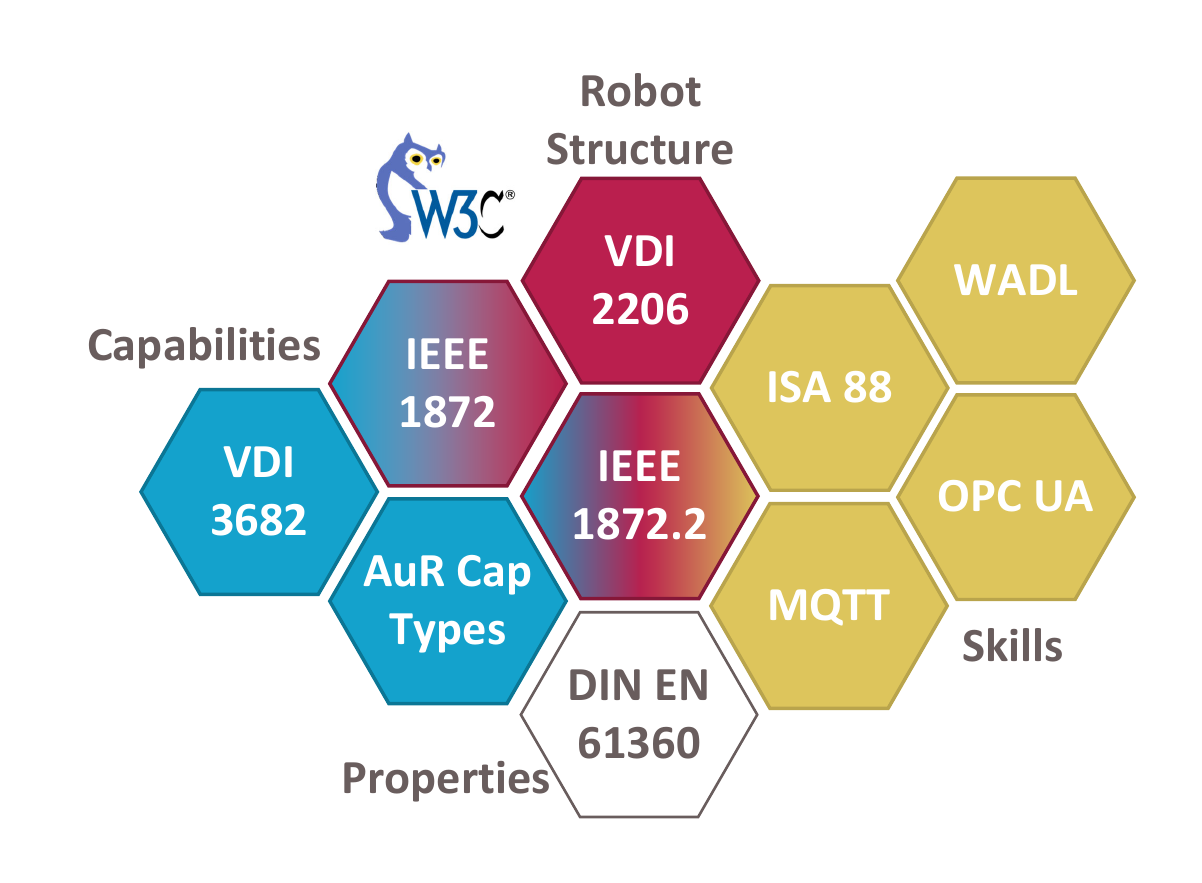}
    \caption{Adapted capability and skill model for heterogeneous autonomous robots.}
    \label{fig:capability-model-aur}
    \end{subfigure}
    \caption{Adaptation of the capability and skill model presented in \cite{Kocher1} to a model suitable for heterogeneous autonomous robots.}
    \label{fig:capability-model-vergleich}
\end{figure*}

Approaches focusing on skills typically use meta models such as AutomationML or OPC UA.
In \cite{BaRe_Digitaldescriptionofproducts_2017}, the authors present an approach to skill-based programming for assembly systems. A skill is defined to be a vendor-neutral description of a machine's functionality that can be invoked by commands. Skills can carry out processes which are in turn needed to assemble a product. The mapping between skills and processes has to be done manually by experts. AutomationML is used to create a model that contains products, processes and skills.

Another approach that focuses on skills is presented in \cite{Profanter.2021}. The approach presents a skill model that is intended to enable a hardware-independent Plug \& Produce system in industrial assembly. For this purpose, OPC UA is used as a standardized skill interface and the skills are described semantically via ontologies, so that composition and execution of skills is possible via a Manufacturing Execution System that uses a knowledge base. Certain capability aspects, e.g., process and product models are also mentioned but no complete capability model is shown. 

In \cite{BBM+_KnowledgeforIntelligentIndustrial_2012} a knowledge-based skill description is also presented. The approach takes place in the domain of work cells. Skills are thereby implemented by a state machine. In the presented ontology, the three aspects product, process and resource are separated from each other, but are connected by skills. A description of the interface for the execution of skills is only partially present in the ontology. 

The model presented in \cite{Kocher1} is the first one to contain both capability as well as skill descriptions in one joint ontology. This ontology consists of multiple so-called ontology design patterns (ODPs) which are all based on industry standards and have been developed and tested in several projects. The ontology shown in \cite{Kocher1} can be divided into three aspects: The \emph{Structure} aspect is used as the basis for capabilities and skills and to capture existing components of machines. The \emph{Capability} aspect is formed by an ontology for formalized process descriptions according to \cite{VDI3682}. Taxonomies of manufacturing and handling operations are further used to type capabilities.
The aspect of \emph{Skills} is centered around a state machine according to \cite{ISA88}. Transitions of this state machine (e.g. starting a skill) may be invoked using either web services or OPC UA. Later publications, e.g., \cite{Kocher2}, extend the model by properties according to \cite{IEC61360}, which enables formal modeling of properties and a clear assignment of properties to resources. In \cite{Kocher2}, a method to automatically generate model instances from PLC code is also added.

In the issue in which this article is published, a first reference model for capabilities, skills and services is also published. This reference model defines the most relevant terms in the field of capabilities, skills and services and shows the essential connections \cite{KBH+_AReferenceModelfor_15.09.2022b}.
A revised version of the model from \cite{Kocher1} covers the aspects of capability and skill of this reference model. As shown in Sec.~\ref{sec:model}, the model presented in this article extends the model of  \cite{Kocher1}, so the presented model also adheres to the reference model.

\subsection{Conclusion on Related Work}
\label{subsec:conclusion}
Based on the approaches presented in Section~\ref{subsec:RW_modelsForRobots} and Section~\ref{subsec:RW_modelsInManufacturing}, it can be shown that on the one hand, in the field of HAuRs, only a few insufficient approaches to function modeling exist. 
On the other hand, there exists a variety of approaches in the area of manufacturing. These approaches use ontologies, which are also proposed in \cite{Bayat.2016} for modeling HAuRs. 
In \cite{Kocher1}, multiple ODPs are created based on standards that can be developed and maintained independently, making the approach of \cite{Kocher1} a good candidate with respect to the requirements of reusability and extensibility. Furthermore, the approach separates capabilities from skills, which is also a requirement for the model of HAuRs. But as this approach is specifically tailored toward manufacturing, it needs adaptions and extensions to be applicable in the field of HAuRs. 
For example, the model does not distinguish autonomous robots in terms of their modality air, land, or water. Likewise, it only considers manufacturing-specific capabilities such as drilling, etc., or does not consider communication or positioning. 
Thus in the following section, extensions to all model aspects of \cite{Kocher1} are presented to derive a model usable for HAuRs.

\section{Capability and Skill Model for Autonomous Robots}
\label{sec:model}
In the capability and skill model presented in \cite{Kocher1}, the aspects of \emph{Structure}, \emph{Capability}, \emph{Property} and \emph{Skill} are considered (cf. Fig.~\ref{fig:capability-model-ak}). These aspects are analyzed and adapted for HAuRs in the following, resulting in the model shown in Fig.~\ref{fig:capability-model-aur}. Each hexagon shows an ODP that is maintained independently and mainly represents an industry standard. An alignment ontology is used to connect individual reusable ontologies. Adopted and new contents of the four aspects are presented, but also links that are established via the alignment ontology.

\begin{figure*}[t]
    \centering
    \includegraphics[width=1.0\linewidth]{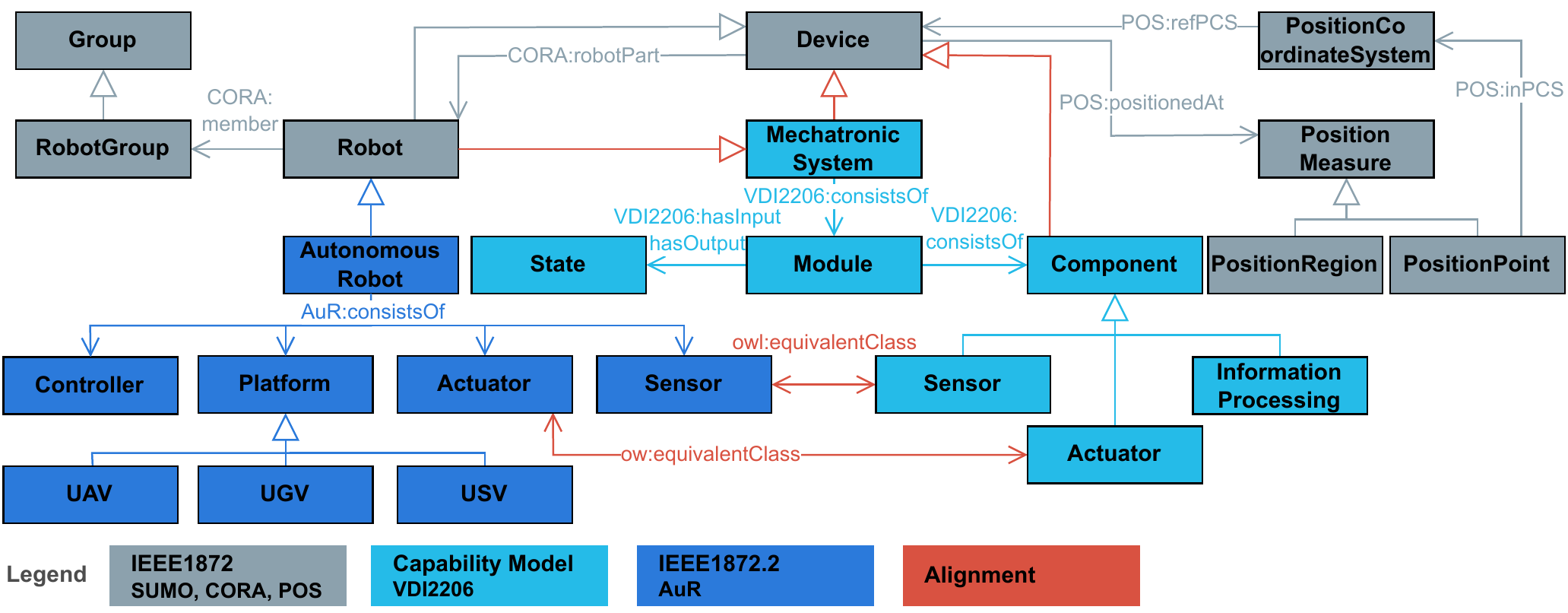}
    \caption{Overview of core classes and relations of the \emph{Structure} aspect of the capability and skill model for heterogeneous autonomous robots.}
    \label{fig:structure}
\end{figure*}

\subsection{Structure}
\label{susbec:structure}
The basis of the capability and skill model for manufacturing is the description of the structure of machines. 
The \emph{Structure} aspect consists of one ODP, which is the industry guideline of VDI 2206 \cite{VDI2206}. 
With \cite{VDI2206}, the defined terms can be used to describe machines as mechatronic systems and their assembly of modules and components, like sensors or actuators. 
HAuRs are also mechatronic systems. Nevertheless, the concepts available in VDI 2206 are too broad and insufficient, e.g., to distinguish different types of HAuRs. 
For the representation of the structure of HAuRs, the IEEE standard 1872.2-2021 \cite{IEEE1872.2} is suitable. Integration with the two upper ontologies SUMO and DUL brings the advantage of a possible connection with other ontologies of other domains. 
Such a connection can be created by subordinating more specific classes to more abstract super classes. Thereby, some concepts of the two upper ontologies have been linked and related to new concepts. In addition, the CORA ontology of the IEEE standard 1872-2015 \cite{IEEE1872} is adopted and connected. The main classes and object properties of the structure of the capability and skill model for HAuRs are shown in Fig.~\ref{fig:structure}. 
In this figure and all the following figures, different colors represent a particular standard that includes one or more ODPs and their existing alignment. The only exception is the additional higher-level alignment to link the different standards. The structure corresponds to a combination of the proven standard of VDI 2206 \cite{VDI2206}, which is applied in the capability and skill model \cite{Kocher1} and adopted, and \cite{IEEE1872.2} including \cite{IEEE1872} and the upper ontologies DUL and SUMO. Fig.~\ref{fig:structure} shows the introduction of a class \emph{Robot} and especially a class \emph{Autonomous Robot} and the distinction of their types. 
\begin{align}
    AutonomousRobot \sqsubseteq Robot \sqsubseteq Device \\
    \exists consistsOf.Platform \sqsubseteq AutonomousRobot 
\end{align}
Linking of the individual ODPs is done via the alignment ontology shown in red. The classes of \cite{VDI2206} have been linked accordingly so that a consistent description is available. For example, \emph{Robot} is a subclass of \emph{Mechatronic System}, allowing \emph{Robot} to also have inputs and outputs. 
\begin{align}
    Robot \sqsubseteq MechatronicSystem 
\end{align}
Due to restricted space, the definition of orientation is not shown, but is consistent with the procedure for defining an individual of class \emph{Position} and thus fulfills the requirement of Sec.~\ref{sec:requirements}. Different devices such as from class \emph{Sensor} can be assigned to a indidivual of class \emph{Robot} via the \emph{robotPart} object property. 
Through this structure, the following statement can be expressed: A drone "Quadrocopter1" is of class \emph{Robot} and consists of components such as rotors and a camera and is in a certain position and orientation.

\subsection{Capabilities}
\label{subsec:caps}
The \emph{Capability} aspect is used to describe functions and make them comparable regardless of different implementations and execution technologies. 
The capability model from manufacturing uses the VDI 3682 \cite{VDI3682} standard for this purpose by defining capabilities as process operators. 
In \cite{VDI3682}, processes consist of multiple process operators, which are described as follows: An individual of class \emph{Process Operator} can be assigned to a individual of class \emph{Technical Resource} and converts a set of inputs into outputs. Inputs and outputs are either of class \emph{Product}, \emph{Information} or \emph{Energy}. 
A \emph{Process Operator} can also be decomposed into further \emph{Process Operators} so that different levels of detail can be described \cite{VDI3682}. 
This approach is adopted in the same way for HAuRs. A capability \emph{transport} can be defined, which has as its input an object to be transported and information about the desired destination. Its output is the object at the desired location. 

\begin{figure*}[t]
    \centering
    \includegraphics[width=1.0\linewidth]{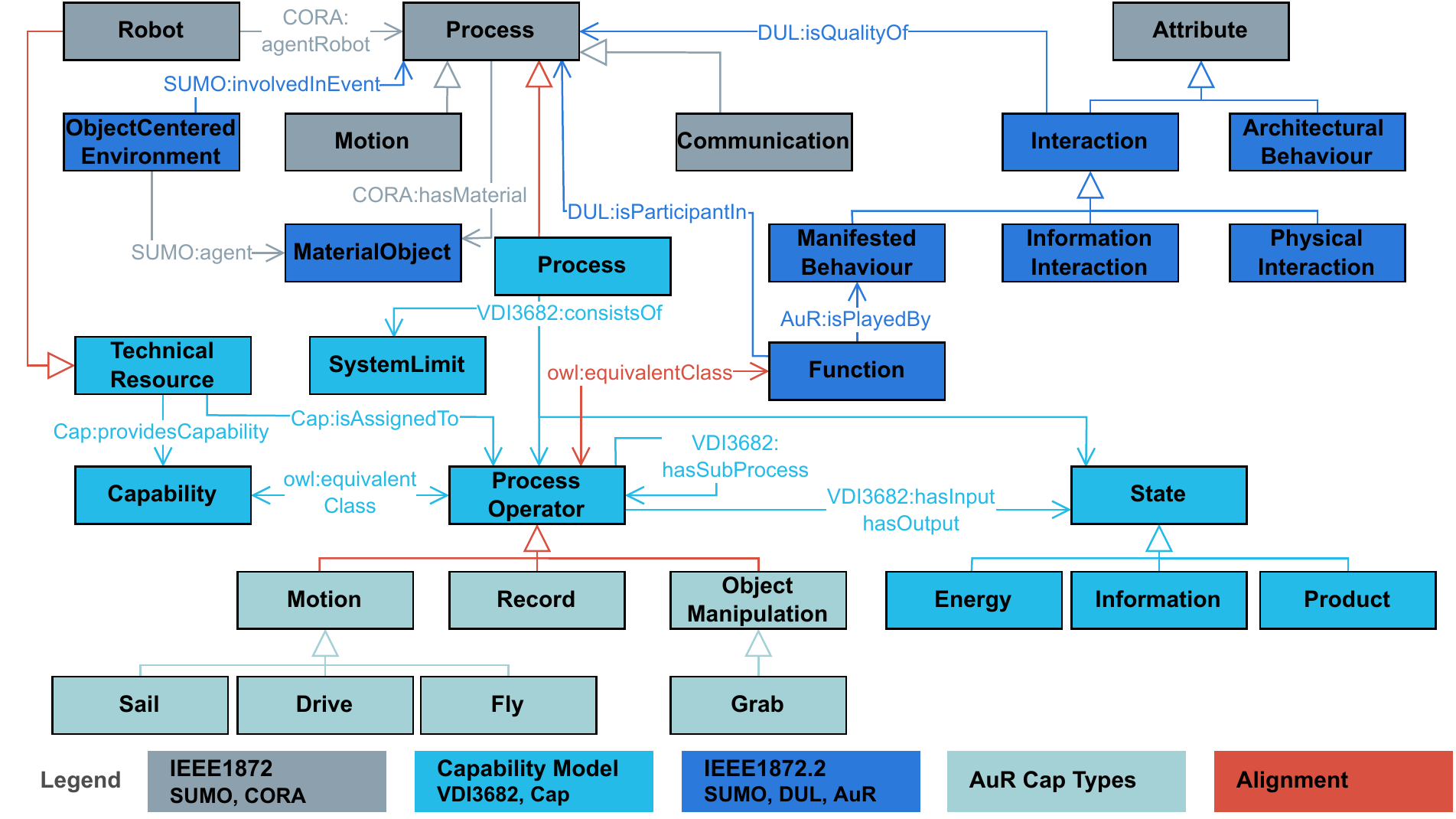}
    \caption{Overview of core classes and relations of the \emph{Capability} aspect of the capability and skill model for heterogeneous autonomous robots.}
    \label{fig:capability}
\end{figure*}

Furthermore, two additional ODPs according to DIN 8580 and VDI 2860 are used in the capability model of \cite{Kocher1}, which define subclasses for the otherwise rather abstract capability class. But these two standards are manufacturing-specific and not relevant to the field of HAuRs. 
Therefore, these ODPs are not considered further. Nevertheless, the approach to give more meaning to the abstract capabilities is to be followed and therefore another ODP is introduced here. However, in the field of HAuRs, there is no standard as in manufacturing where, e.g., manufacturing processes are described. Therefore, a separate ODP with suitable classes of capabilities has to be created.

Fig.~\ref{fig:capability} gives an overview of the aspect of capabilities for HAuRs. 
Every \emph{Capability} is modeled as a \emph{Process Operator}, so the two classes are equivalent. Every \emph{Capability} is provided by \emph{Technical Resource} via \emph{providesCapability}. The two introduced standards of the IEEE are also applied here, since the classes \emph{Interaction}, \emph{Communication} as well as \emph{Environment} are considered, which were defined as requirements in Sec.~\ref{sec:requirements}. 
The class \emph{Process} considers only a few capability types such as \emph{Motion} and \emph{Communication}, so more specific capability types such as grasping an object are represented via the additional ODP, here called \emph{AuR-Cap}. 
The specific capability types are subclasses of \emph{Process Operator} to assign a capability to a specific type. To connect all these ODPs, some alignments have been made, e.g., a subclassification of \emph{Process} from \cite{VDI3682} to \emph{Process} from \cite{IEEE1872}. 
\begin{align}
    VDI3682\!:\!Process \sqsubseteq SUMO\!:\!Process
\end{align}
In \cite{IEEE1872.2} a separation of functions from the execution of functions is also addressed. Therefore, \emph{Process Operator} is aligned with the corresponding \emph{Function} class. A \emph{Robot} is subclass of \emph{Technical Resource}, so a \emph{Robot} can be assigned to a \emph{Capability}. Accordingly, the connections of cross-domain ODPs are established.   
\begin{align}
    ProcessOperator \equiv Function \\
    Robot \sqsubseteq TechnicalResource 
\end{align}

\subsection{Properties}
\label{subsec:props}
The formal description of properties was introduced previously in the capability model for manufacturing. In the capability model, the concepts defined in \cite{IEC61360} are used to describe properties \cite{Kocher2}. The concepts of \cite{IEC61360} offer a generic possibility to describe different properties formally. 
Each object can have properties assigned to it. Each property, such as height, is uniquely defined once in the model via an individual of so-called class \emph{TypeDescription}. 
With a \emph{TypeDescription}, information like a textual definition or the unit of measure are fixed.
Objects sharing properties of the same \emph{TypeDescription} all need to have an individual of class \emph{DataElement} related to this \emph{TypeDescription}. Specific occurences of a property are modeled using so-called \emph{InstanceDescriptions}, which are also linked to the corresponding \emph{DataElement}. 
\begin{figure*}[t]
    \centering
    \includegraphics[width=1.0\linewidth]{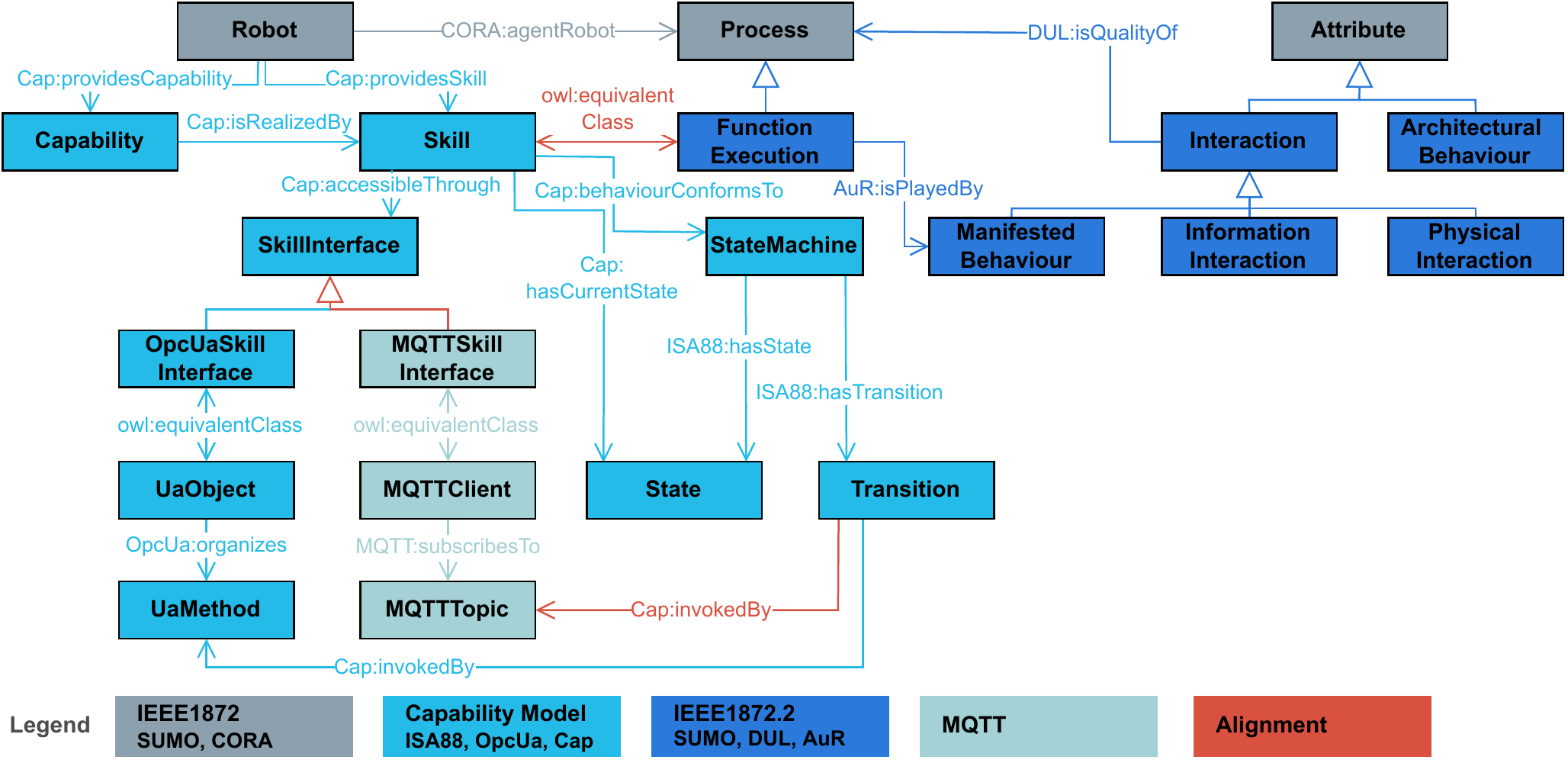}
    \caption{Overview of core classes and relations of the \emph{Skill} aspect of the capability and skill model for heterogeneous autonomous robots.}
    \label{fig:skill}
\end{figure*}
An \emph{InstanceDescription} can be used to specify an actual value, a requirement, or an assurance about a value or range of values. Multiple \emph{InstanceDescriptions} can be attached to a single \emph{DataElement}. This machine-readable description of properties is also applicable to HAuRs. It allows to formulate constraints on the capabilities that certain robots provide. An example of an assurance is: a drone "Hexacopter2"has the ability to fly up to an altitude of 100 meters above ground.

\subsection{Skills}
\label{subsec:skills}
The fourth aspect is devoted to skills and comprises the description of interfaces that can be used to execute them. A central part of this aspect in \cite{Kocher1} is the state machine standardized in ISA 88 \cite{ISA88}. 
It is used to describe the states and transitions of a skill. On this basis, the current state of a skill is tracked on the one hand, and permissible transitions of a skill are described on the other hand. This creates a technology-agnostic interface to interact with skills. 
But in order to be callable, transitions must be mapped to execution technologies. In \cite{Kocher1}, this is achieved with two other ODPs for (a) RESTful webservices described using Web Application Description Language (WADL) and (b) OPC UA. 
The WADL ODP enables the description of web services in order to be able to access them via a given URL using HTTP methods. The OPC UA ODP can be used to describe an OPC UA Server and its objects with methods and variables according to the skills, their transitions and parameters.

These three ODPs can also be used for HAuRs. In \cite{IEEE1872.2} executable functions are also separated from functions. \emph{Function Execution} is modeled to be equivalent to the class \emph{Skill}. 
\begin{align}
    Skill \equiv FunctionExecution 
\end{align}
Another technology that is applied in the field of HAuRs is MQTT \footnote{https://mqtt.org/}. MQTT is a standard messaging protocol for the Internet of Things (IoT), which can also be ideally used for robot-to-robot communication. 
MQTT uses a publish/subscribe pattern, which differs from classic client-server architectures in that a third party, a broker, is interposed. Publishers are clients that send a message and subscribers are clients that receive messages. Publishers and subscribers are decoupled from each other by the broker and do not know each other. The broker filters incoming messages and distributes them to the clients. The broker manages so-called topics to which clients can publish or subscribe \cite{Soni.2017}. Accordingly, an \emph{MQTT Client} is a \emph{SkillInterface} that listens for certain topics that can be reached via the broker. 
\begin{align}
    MQTTClient \equiv MQTTSkillInterface \\
    MQTTSkillInterface \sqsubseteq SkillInterface
\end{align}
An individual of class \emph{Transition} of an indiviual of class \emph{StateMachine} is triggered by publishing a message to the corresponding \emph{MQTTTopic}, e.g. to start the skill. Basically, the description is done in a similar way as for WADL or OPC UA. Fig.~\ref{fig:skill} provides an overview of the core concepts. 
An individual of class \emph{Skill} represents the executability of a \emph{Capability} via \emph{isRealizedBy} and is \emph{accessibleThrough} an \emph{SkillInterface} which has different technological variants as subclasses, thus connecting these Skill ODPs to the ontology. A Skill has a current \emph{State} and a \emph{StateMachine} from which a particular \emph{Transition} is executable by a particular method specific to the execution technology. 

\section{Conclusion}
\label{sec:conclusion}
The paper presented a formal capability and skill model for HAuRs. For this purpose, existing approaches to describe HAuRs and their functions have been considered. So far, there are few and insufficient approaches in the field of HAuRs for function modeling. Since the topic of function modeling is already addressed in the field of manufacturing, approaches from this field were considered. Based on requirements for modeling HAuRs and their functions as summarized in this paper, it was shown that models from manufacturing are promising but still lack some aspects for application to HAuRs. 

Therefore, an existing capability model from manufacturing that most closely matches the requirements was built upon and extended. The model aims to describe HAuRs and their functions, both abstractly in terms of capabilities and technology-dependent in terms of skills. 
Extensions were made in the form of ODPs, which are based on general or domain standards, both to represent the structure of an autonomous robot and to add robot-specific capability types. The extensions allow the description of positions, interaction and environment aspects. 
On the executable skills side, MQTT technology was added as an skill interface. The use of ODPs also enables future extensions of the capability and skill model. The presented model can be found at \url{https://github.com/Miguel2617/robocap}. 

The model allows the description of HAuRs in a formal way, so that collaboration between these robots is easily enabled. In the future, this collaboration will be shown in an implementation that uses the capability and skill model and thus allows the registration of robots and their capabilities but also the execution of skills via skill interfaces.

Ideally, the creation of the information model should be integrated into the development of a robot, so that parallel to a robot also its capability and skill model is created. With proper tool support, this could even be achieved without the need for ontology expertise. We aim to develop an engineering method with corresponding tool support in future research. 
In addition, future research will also be devoted to methods for the automated creation of descriptions from various artifacts such as describing the structure from Unified Robot Description Format (URDF) documents of a robot.

The separation of abstract capabilities and executable skills enables mission planning at the abstract capability level and mission execution at the skill level. A team of HAuRs can be assigned a task, which can be reviewed based on the offered abstract capabilities and planned if the task can be executed. 

The described approach to develop an ontology based on existing ODPs and employing domain-specific standards can serve as a blueprint for engineers in other domains who aim at a machine-interpretable description of the capabilities and skill interfaces of their system's components.   

\begin{acknowledgements}
  \noindent This research is funded by the project RIVA as part of dtec.bw – Digitalization and Technology Research Center of the Bundeswehr, which we gratefully acknowledge.
\end{acknowledgements}

\bibliographystyle{IEEEtran}
\bibliography{bib/references.bib}
\end{document}